\documentclass{article}
\usepackage{spconf,amsmath,graphicx}

\usepackage{pstricks,graphicx}
\usepackage{setspace}
\usepackage{psfrag}

\usepackage{amsmath,graphicx}

\usepackage{pstricks,graphicx}
\usepackage{setspace}
\usepackage{psfrag}
\usepackage{amssymb}
\usepackage{amsmath}
\usepackage{amsthm}

\title{ %{\footnotesize \emph{newly accepted by IEEE International Conference on Acoustics, Speech, and Signal Processing (ICASSP), 2013}}
Object Detection and 3D Estimation via an FMCW Radar Using \\ a Fully Convolutional Network}
\graphicspath{{figures/}}

\name{Guoqiang Zhang$^*$, Haopeng Li$^\dagger$, and Fabian Wenger$^\dagger$ }
\address{$^*$ University of Technology Sydney, Australia \\
$^\dagger$  Qamcom Research and Technology AB, Sweden 
\thanks{Author emails: guoqiang.zhang@uts.edu.au, haopeng.li@qamcom.se, fabian.wenger@qamcom.se
}
}

\begin{document}
\ninept

%\onecolumn

\maketitle
\begin{abstract}
This paper considers object detection and 3D estimation using an FMCW radar. The state-of-the-art deep learning framework is employed instead of using traditional signal processing. In preparing the radar training data, the ground truth of an object orientation in 3D space is provided by conducting image analysis, of which the images are obtained through a coupled camera to the radar device. To ensure successful training of a fully convolutional network (FCN),  we propose a normalization method, which is found to be essential to be applied to the radar signal before feeding into the neural network. The system after proper training is able to first detect the presence of an object in an environment. If it does, the system then further produces an estimation of its 3D position. Experimental results show that the proposed system can be successfully trained and employed for detecting a car and further estimating its 3D position in a noisy environment.

\end{abstract}

\begin{keywords}
FMCW radar, camera, U-Net, FCN, object detection.
\end{keywords}

\vspace{-2mm}
\section{Introduction}
\vspace{-1mm}

Reliable object detection using one or more sensors is critical for applications like autonomous driving \cite{Levinson11AutoDrive}, interactive video games, and surveillance tasks.  Typical sensors for object detection include cameras, radars, and LiDARs.  In general, different sensors have their unique sensing properties, which brings each type of sensor an advantage over others when performing object detection. For instance, cameras are able to capture rich texture information of objects in normal light conditions, which makes it possible to identify and distinguish objects from background. Radars attempt to detect objects by continuously transmitting microwaves and then analyzing the received signals reflected by the objects,  which allow the sensors to work regardless of bad weather conditions or dark environments. 

In recent years, object detection based on cameras has made significant progress by using deep learning framework. The basic idea is to design and train a deep neural network (DNN) by feeding a large number of annotated image samples. The training process enables the DNN to effectively capture informative image features of interested objects via multiple neural layers  \cite{Lecun15nature}. As a result, the trained DNN is able to produce impressive performance for visual object detection and other similar tasks such as object classification and segmentation (e.g.,  Mask R-CNN \cite{He17MaskRCNN},  YOLO  \cite{Redmon16Yolo}, and U-Net \cite{Ronneberge15UNet}).

\begin{figure}[t!]
\centering
\includegraphics[width=70mm]{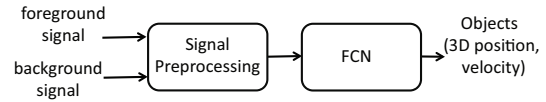}
\caption{\small {Diagram of the proposed object detection and 3D estimation system via an FMCW radar at a fixed place using an FCN. The background radar signal only contains reflected noise introduced by the environment.  Information of interested objects is only embeded in the foreground signal.  The FCN exploited in this work is a variant of U-Net.} }
\label{fig:diagram}
\vspace{-3mm}
\end{figure}

\begin{figure}[t!]
\centering
\includegraphics[width=80mm]{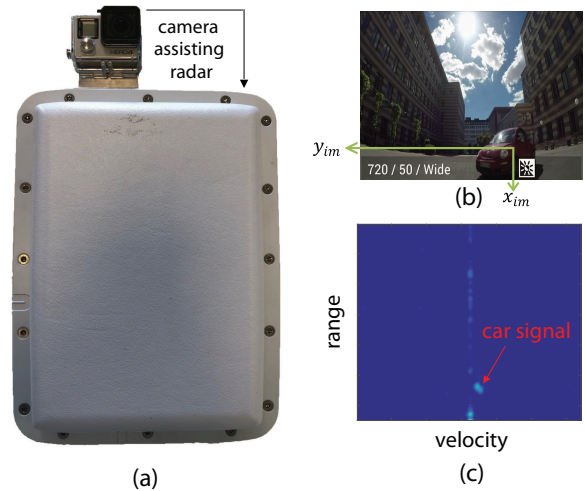}
\caption{\small (a):  radar (QR77SAW from Qamcom Research and Technology AB) plus a coupled camera; (b) camera image; (c) range-doppler spectrum from one antenna receiver. The camera assists the radar by annotating the radar signals to allow for FCN training. The image coordinates $(x_{im},y_{im})$ are firstly estimated through image analysis, and then treated as the ground truth of the object orientation when training the FCN for analyzing the radar signal. }
\label{fig:RadCam}
\vspace{-5mm}
\end{figure}

Research on exploiting DNNs for analyzing radar signals is still at an early stage. \cite{Capobianco17Radar} considered the problem of classifying 6 different vehicles using the frequency-modulated continuous-wave (FMCW) radar signals, where Short Time Fourier Transformation (STFT) is firstly applied to the original radar signals to obtain spectrums as inputs to the DNN. In \cite{Micka15Radar}, the authors attempted to detect the presence of vehicles using DNNs, which can be formulated as a binary classification problem. The work of \cite{Hadhrami18RadarCNN} considered combining DNNs and support vector machine (SVM) for moving radar target classification.  The above classification tasks do not fully exploit the information embedded in radar signals for advanced object detection such as range and velocity estimation of interested objects. To the best of our knowledge, there is \emph{no prior work on using DNNs to simultaneously detect the presence and estimate the 3D positions of objects (e.g., vehicles) based on radar signals}.

In this work, we attempt to fully exploit the FWMC radar signals to detect the presence and estimate the 3D positions of objects based on DNNs. It is known that for an FWMC radar with multiple antenna receivers, 3D information (i.e., range, elevation and azimuth) of interested objects is embedded in the received radar signals \cite{Nam18Radar, Laribi17Radar}. Our motivation for exploiting the DNN-based approach is that radar signals can be preprocessed and treated as images. By doing so, the obtained knowledge of employing DNNs for successful image analysis in the literature could be transferred to radar signal analysis. %It is known that DNNs have achieved great success in many vision-related problems (see \cite{Lecun15nature}).

The new DNN-based system is designed by following the diagram in Fig.~\ref{fig:diagram}, which consists of a \emph{signal preprocessing} block and a fully convolutional network (FCN) block. A background radar signal is processed together with a foreground signal to be able to combat reflection noises introduced by the environment. The proposed system aims to detect and estimate 3D information of one object only appearing in the foreground. 

In brief, we make three contributions towards successful usage of an FCN for reliable radar-based object detection and 3D estimation. Firstly, in preparation of training data, we use a coupled camera to annotate radar signals (see Fig.~\ref{fig:RadCam}).  Suppose the radar training signal is for estimating the range, azimuth and elevation of one object. The ground truth of azimuth and elevation will be provided by conducting image analysis, assuming that the radar signal and the corresponding image sequence are well synchronized.  

Secondly, we propose a normalization method for radar signal which works together with 2D-FFT as the preprocessing block for the  system in Fig.~\ref{fig:diagram}. Suppose a foreground (or background) radar signal segment is transformed to $N$ range-doppler spectrums after 2D-FFT, one for each radar receiver. The normalization method operates on each range-doppler cell of the $N$ spectrums to cancel out the effect of phase shift of radar signals due to range-difference in space. The normalization is essential to ensuring successful training of the FCN later on. 
  
%by taking the data of the first receiver as a benchmark.  By doing so, the normalization method  removes ambiguities caused by the phase-shift, ensuring successful training of the FCN in the 2nd step. 

Thirdly, we propose a variant of U-Net (one type of FCN \cite{Ronneberge15UNet}) to analyze the normalized range-doppler spectrums obtained from the signal preprocessing block. The proposed network firstly detect presence of objects in the foreground. If an object is identified, the network then further estimates its azimuth and elevation to fully determine its 3D location.  %which is formulated as a regression problem at the final neural layer. 
As an example, we successfully trained the radar system for detecting and estimating the 3D position of a car in  a noisy environment.

\vspace{-2mm}
\section{Preliminary}
\vspace{-1mm}

In this section, we briefly explain how the 3D information of an object is embedded in the  radar signals of an FMCW radar with $N$ receivers. The difference between range-doppler spectrums of radar signals and camera images will also be briefly discussed.

Suppose an FMCW radar keeps transmitting a frequency modulated microwave signal in its front field.  A stationary object in the field would reflect back the signal to the radar device, which is actually a delayed and damped version of the transmitted signal. Information of the range or distance between the radar and the object is naturally embedded in the time delay. Considering a moving object in the field, the delay would vary over time if the object has nonzero radial velocity w.r.t. the radar device. In principle, the radial velocity should be able to be computed by measuring the delay change over time \cite{Nam18Radar, Laribi17Radar}.          

\begin{figure}[t!]
\centering
\includegraphics[width=65mm]{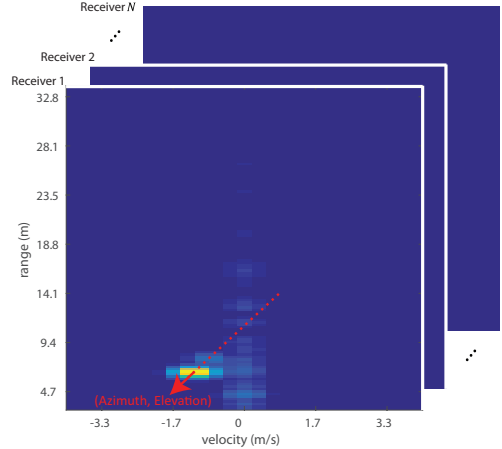}
\caption{\small $N$ range-doppler spectrums of an FMCW radar with $N$ receivers, one spectrum for each receiver. Information of azimuth and elevation of an object is embedded in the corresponding range-doppler cell of the $N$ spectrums.  }
\label{fig:spectrum}
\vspace{-5mm}
\end{figure}

It is found that the range and radial velocity of an object corresponds to the vertical and horizontal axis of the spectrum obtained by performing 2D FFT on a radar signal segment \cite{Nam18Radar, Laribi17Radar}, which is usually referred to as the \emph{range-doppler spectrum}.  As shown for the ideal case in Fig.~\ref{fig:spectrum}, the range and radial velocity of an object can be easily obtained by searching for the coordinates of the highest signal magnitude in the spectrum. In practice, a noisy environment might cause the object signal be masked by background noise, making it challenging to obtain an accurate estimation. 

Next we consider estimating the object orientation in the form of azimuth and elevation. Suppose the radar device has $N$ antenna receivers, which are properly distributed inside its radome. Depending on the orientation of the object w.r.t. the radar device, the reflected radar signal from the object would arrive at the $N$ receivers with different time patterns. Therefore, the different time-of-arrivals (TOAs) carry the azimuth and elevation information of the object. After obtaining $N$ range-doppler spectrums (one for each receiver), information of the object orientation is naturally embedded in phase domain of the spectrum (see Fig.~\ref{fig:spectrum} for demonstration).  
 
 As will be explained later on,  spectrums of radar signals will be treated as images to allow for using the FCN-based image analysis framework in the literature.  While the pixel position from a camera image roughly represents the orientation of an object in 3D space, the cell position of radar spectrums represents the range and radial velocity of an object. Furthermore, the azimuth and elevation information of an object is carried in the phase domain of the corresponding range-doppler cells over $N$ receivers. In brief, radar spectrums are fundamentally different from camera images. Each signal type provides a unique set of features which may benefit the other in certain applications.     

 \begin{figure*}[t!]
\centering
\includegraphics[width=150mm]{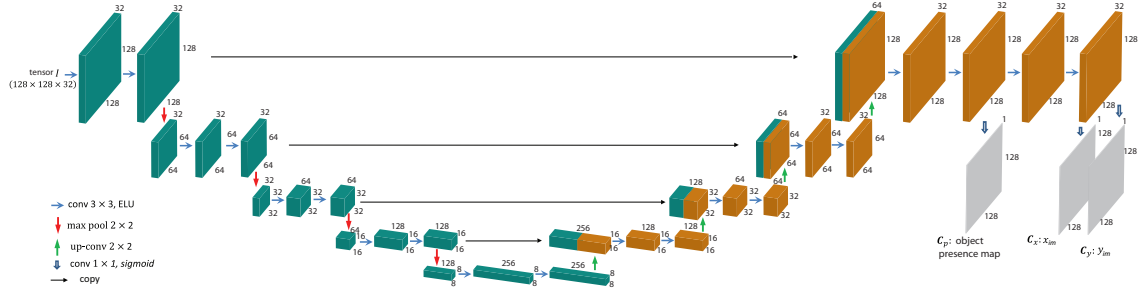}
\caption{\small The FCN structure, which is a variant of U-Net. The input tensor $I$ includes information of both foreground  and background radar signals. The neural network produces three outputs: the object presence map $\boldsymbol{C}_p$, and the two maps $\boldsymbol{C}_{x}$ and $\boldsymbol{C}_y$ for estimating the image coordinates $(x_{im},y_{im})$ of the object.}
\label{fig:UNet}
\vspace{-3mm}
\end{figure*}

\vspace{-2mm}
\section{On radar signal annotation using \\ a coupled camera}
\vspace{-1mm}
\label{sec:annotation}

Radar signal annotation is the key step to allow for the FCN training in the later stage. To do so, we need to provide the ground truth of 3D position (i.e., range, azimuth $\varphi $ and elevation $\theta$ ) of an object as well as its cell location (see Fig. \ref{fig:spectrum}) in the range-doppler spectrums. The range and cell location can be simultaneously obtained by manually marking the range-doppler spectrums. It is challenging to acquire the ground truth of the azimuth $\varphi $ and elevation $\theta$ of the object by using the radar device alone. 

To facilitate radar signal annotation, we propose a novel solution to obtain the ground truth for the orientation of an object. As shown in Fig.~\ref{fig:RadCam}, we propose to use a coupled camera of the radar device to estimate the orientation of the object.  It is known that under good light conditions, image analysis can often provide an accurate estimation of the image coordinates $(x_{im},y_{im})$ of the object (see Fig.~\ref{fig:RadCam} (b) for demonstration). Suppose the camera is fixed w.r.t. the radar device, it is straightforward that the image coordinates $(x_{im},y_{im})$ hold a one-to-one mapping to $(\varphi, \theta)$. If the coordinates of the radar and the camera are probably calibrated, $(\varphi, \theta)$ can then be easily computed from $(x_{im},y_{im})$, which can then be taken as the ground truth for the FCN training later on. 

Radar-camera calibration is usually time consuming and requires special equipments and computing programs. In this work, we avoid the step of radar-camera calibration.  Instead, the image coordinates $(x_{im},y_{im})$ of the object is taken directly as the ground truth of the object orientation. The FCN in Fig.~\ref{fig:diagram} is designed to predict $(x_{im},y_{im})$ of the object directly instead of $(\varphi, \theta)$. 

Our motivation for estimating the image coordinates $(x_{im},y_{im})$ instead of $(\varphi,\theta)$ is based on the hypothesis that the FCN would be able to implicitly learn the coordinate-mapping between camera and radar. As will be discussed in Section \ref{equ:experiment}, the experimental results justify our hypothesis nicely.  

The ability of estimating the image coordinates $(x_{im},y_{im})$ directly from the neural network makes our system simple and practical. Firstly, there is no need to calibrate the radar and camera w.r.t. a common coordinate system. The range and image coordinates together are able to determine 3D position of an object.  Secondly, it simplifies the annotation procedure of radar training samples using the coupled camera. Once the image coordinates of an object are obtained using the camera system, they will be used directly to label the training samples. 

\vspace{-2mm}
\section{On Signal Preprocessing and FCN Training}
\vspace{-1mm}

\subsection{System description}
\begin{figure}[t!]
\centering
\includegraphics[width=70mm]{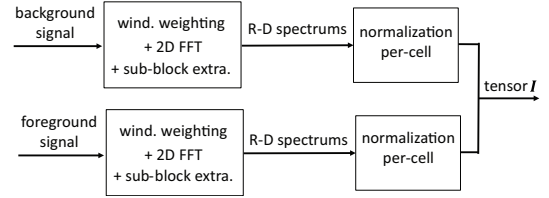}
\caption{\small Elaboration of the signal preprocessing block in Fig.~\ref{fig:diagram}. }
\label{fig:prePro}
\vspace{-5mm}
\end{figure}

As depicted in Fig.~\ref{fig:diagram}, the proposed system consists of two blocks. The first block performs preprocessing to both a background and foreground time-domain radar signal segments. The background signal only contains noise from the environment. It is introduced to assist the system in detecting an object that only appears in the foreground. As shown in Fig. \ref{fig:prePro}, the first block includes two basic operations which are 2D FFT and phase-normalization per range-doppler cell.  After the two operations, each segment yields $N$ normalized range-doppler spectrums  in the complex domain, one for each radar receiver.  In total, there are $2N$ normalized range-doppler spectrums.  

The second block is an FCN to further analyze the $2N$ spectrums and perform object detection and 3D estimation. In particular, it first detects the presence of an object in the foreground. If an object is identified in the foreground, the neural network further estimates the range, and the image coordinates $(x_{im}, y_{im})$ in the image of the coupled camera. 

%Differently from traditional signal processing,  we first have to properly train the FCN using a reasonable number of annotated training samples before exploiting it for practical usage. In preparation of the training samples, we make use a coupled camera to annotate the radar signals by providing estimation of $(\varphi, \theta)$ for the objects appearing in foreground.  

\vspace{-3mm}
\subsection{Phase-normalization}
In this subsection, we present the phase-normalization step on the obtained range-doppler spectrums as shown in Fig.~\ref{fig:prePro}. We first briefly clarify the approximate independence between object distance and orientation. Suppose an object is at the far field of the radar device, where the object distance is significantly larger than the microwave length sent out by the radar device. In this case, the object orientation $(\varphi, \theta)$ is (roughly) independent of the object distance. That is, if the object straightly moves either towards to or away from the radar device, its orientation $(\varphi, \theta)$ remains roughly the same. 

The above analysis suggests that one can freely multiply a rotation scalar (i..e, $e^{j \phi}$ for any $\phi\in \mathbb{R}$) to each range-doppler cell across the $N$ spectrums without affecting the object orientation.  Therefore, in our work, we normalize the phases of each range-doppler cell (corresponding to an $N$ dimensional vector) by taking the spectrum of the first receiver as a benchmark. After normalization, the spectrum of the first receiver always has zero phases. 

We note that the normalization step is crucial to successfully train the FCN and further utilize the network for object detection and 3D estimation. With normalization, the FCN does not need to figure out by itself that the object range is unrelated with the estimation of object orientation,  making the training process feasible.    

% The operation cancels out the effect of phase shift of radar signals due to range-difference in space by taking the data of the first receiver as a benchmark.  The normalization method essentially removes ambiguities caused by the phase-shift, ensuring successful training of the FCN in the 2nd step.

\vspace{-3mm}
\subsection{FCN architecture and loss function}

\subsubsection{Structure of the neural network} We exploit an FCN to analyze the tensor $\boldsymbol{I}$ obtained from the signal preprocessing step. Fig.~\ref{fig:UNet} displays the variant U-Net (one type of FCN) exploited in our work.  In total, it has 28 hidden layers and three outputs.  The 28 hidden layers include 20 conv. layers,  4 max-pooling and 4 up-conv. layers.  The first output is the object presence map, which we denote as $\boldsymbol{C}_p$. Each cell variable  $\boldsymbol{C}_p(k,m)$ represents a binary probability, indicating the likelihood of an object occupying the cell $(k,m)$. The second and third outputs represent the estimates of object orientation in terms of $x_{im}$ and $y_{im}$,  which we denote as $\boldsymbol{C}_x$  and $\boldsymbol{C}_y$.  Correspondingly, the two cell variables $\boldsymbol{C}_x(k,m)$  and $\boldsymbol{C}_y(k,m)$ represent the estimate of $x_{im}$ and $y_{im}$ of the object at cell $(k,m)$ if it exists.  

%As shown in Fig.~\ref{fig:UNet}, two  conv. layers are inserted between the two output layers  $\boldsymbol{C}_p$ and $(\boldsymbol{C}_x,\boldsymbol{C}_y)$.  The motivation for introducing the two layers is to bring sufficient expressive power to the neural network so that the image coordinates  $(\boldsymbol{C}_x,\boldsymbol{C}_y)$ of the objects in the foreground can be estimated with high accuracy.  As they are conv. layers, only a limited number of parameters are introduced which is less likely to incur overfitting.
 
\subsubsection{Loss function} So far the FCN structure has been motivated and explained. We now  briefly describe the loss function needed for training the FCN. As analyzed from above, the first output $\boldsymbol{C}_p$ of the neural network estimates object presence in the foreground, which is equivalent to an image segmentation problem (see \cite{Ronneberge15UNet}). We therefore design the loss function for $\boldsymbol{C}_p$ to be a combination of binary cross-entropy and a Dice loss \cite{Milletari16VNet}, denoted as $f_{seg}(\boldsymbol{C}_{p},\boldsymbol{C}_{p}^g)$, where $\boldsymbol{C}_{p}^g$ represents the ground truth. The second output $(\boldsymbol{C}_x,\boldsymbol{C}_y)$ further determines the object orientation detected in the first output by providing estimates of their image coordinates. 
We therefore measure the mean squared error (MSE) between the estimates $(\boldsymbol{C}_x,\boldsymbol{C}_y)$ and their ground truth $(\boldsymbol{C}_x^g,\boldsymbol{C}_y^g)$, denoted as $\|\boldsymbol{C}_{x}- \boldsymbol{C}_{x}^g \|^2$  and   $\|\boldsymbol{C}_{y}- \boldsymbol{C}_{y}^g \|^2$, respectively. When training the FCN, a summation of the above three losses is minimized through backpropogation.       

\vspace{-2mm}
\section{Experiments}
\vspace{-1mm}
\label{equ:experiment}

In the experiment, the radar QR77SAW from Qamcom Research and Technology AB was employed for evaluating the proposed object detection and 3D estimation system. The radar has one transmitter and $N=8$ receivers.  As shown in Fig.~\ref{fig:RadCam}, a camera was mounted at the top of the radar for both radar signal annotation and detection visualization. The radar signal and image sequences from the camera were properly synchronized as required by the proposed system.

The experiment was designed for the radar to detect and estimate the 3D position of a car in an environment with surrounded buildings as shown in Fig.~\ref{fig:RadCam}. The tested range for the car was between 4 m to 28 m. Three segments of radar and camera data were collected separately: one for the background (i.e., no car in the environment) and the other two for the foreground (i.e., a car moving in the field). The background segment contains 800 radar-camera frames while the first and second foreground segments have 2214 and 2323 frames, respectively.  As the radar was placed by facing the front ground surface rather than sky,  strong background noise exists in the collected radar signal.  

In preparation for evaluating our system, all the foreground radar-camera frames were carefully annotated by following the guidelines in Section \ref{sec:annotation}. That is, the ground truth for the car orientations in the radar signal were obtained by estimating the image coordinates of the car through image analysis, which is further manually checked for correctness. The obtained image coordinates $(x_{im}, y_{im})$ were normalized to the range $[0,1]$ to facilitate FCN training. The cell positions of the car in the range-doppler spectrums were manually marked. 

The first foreground segment was selected for training the FCN while the second one was for performance validation. In particular, 2214 training samples were generated by randomly pairing the frames from the first foreground segment and the background frames. Similarly, 2323 validation samples were generated by using the background and the second foreground segments. The stochastic gradient decent (SGD) method was chosen for training the FCN, of which the learning rate and momentum were set to 0.03 and 0.9, respectively. In total, the neural network was trained for 200 epochs from scratch. 

The training results were briefly summarized in Table~\ref{tab:loss_2nd}. It is seen that the MSE for $\boldsymbol{C}_x$ is slightly larger than that for  $\boldsymbol{C}_y$. This is because when collecting the data,  the car moved on the ground surface in a horizontal manner. As a result, the coordinate $y_{im}$ was always within a small range while the coordinate $x_{im}$ changed a lot as the car moved. One observes that the validation loss for $\boldsymbol{C}_p$ is noticeably larger than the training loss compared to those for   $\boldsymbol{C}_x$ and  $\boldsymbol{C}_y$. This might be due to the fact that the segmentation problem for $\boldsymbol{C}_p$ is difficult to train compared with the regression problems for $\boldsymbol{C}_x$ and $\boldsymbol{C}_y$ in our system.  %. the collected radar signal is noisy and it is not a big dataset.  %which makes the validation data has slightly different statistics from the training dataset.   

%\begin{table}[h]
%\caption{\small List of training and validation losses after 1000 epochs. } 
%\label{tab:loss}
%\centering
%\begin{tabular}{|c|c|c|c|c|c|c|c|}
%\hline
% & {\footnotesize loss for $\boldsymbol{C}_p$}&{\footnotesize MSE for $\boldsymbol{C}_x$ }
%&  { \footnotesize MSE for $\boldsymbol{C}_y$} \\
%\hline 
%{\footnotesize \textrm{training}} & {\footnotesize -0.76}  & {\footnotesize $3.0\times 10^{-3}$} & {\footnotesize $8.0\times 10^{-5}$}  \\
%\hline
%{\footnotesize \textrm{validation}} & {\footnotesize -0.48}  & {\footnotesize $4.0\times 10^{-3}$} &  {\footnotesize $9.0\times 10^{-5}$} \\
%\hline 
%\end{tabular}
%\vspace{-2mm}
%\end{table}
\vspace{-4.5mm}

\begin{table}[h]
\caption{\small List of training and validation losses after 200 epochs. } 
\label{tab:loss_2nd}
\centering
\begin{tabular}{|c|c|c|c|c|c|c|c|}
\hline
 & {\footnotesize loss for $\boldsymbol{C}_p$}&{\footnotesize MSE for $\boldsymbol{C}_x$ }
&  { \footnotesize MSE for $\boldsymbol{C}_y$} \\
\hline 
{\footnotesize \textrm{training}} & {\footnotesize -0.63}  & {\footnotesize $2.8\times 10^{-3}$} & {\footnotesize $7.7\times 10^{-5}$}  \\
\hline
{\footnotesize \textrm{validation}} & {\footnotesize -0.52}  & {\footnotesize $3.9\times 10^{-3}$} &  {\footnotesize $8.2\times 10^{-5}$} \\
\hline 
\end{tabular}
\vspace{-2mm}
\end{table}

\begin{figure}[t!]
\centering
\vspace{-5mm}
\includegraphics[width=70mm]{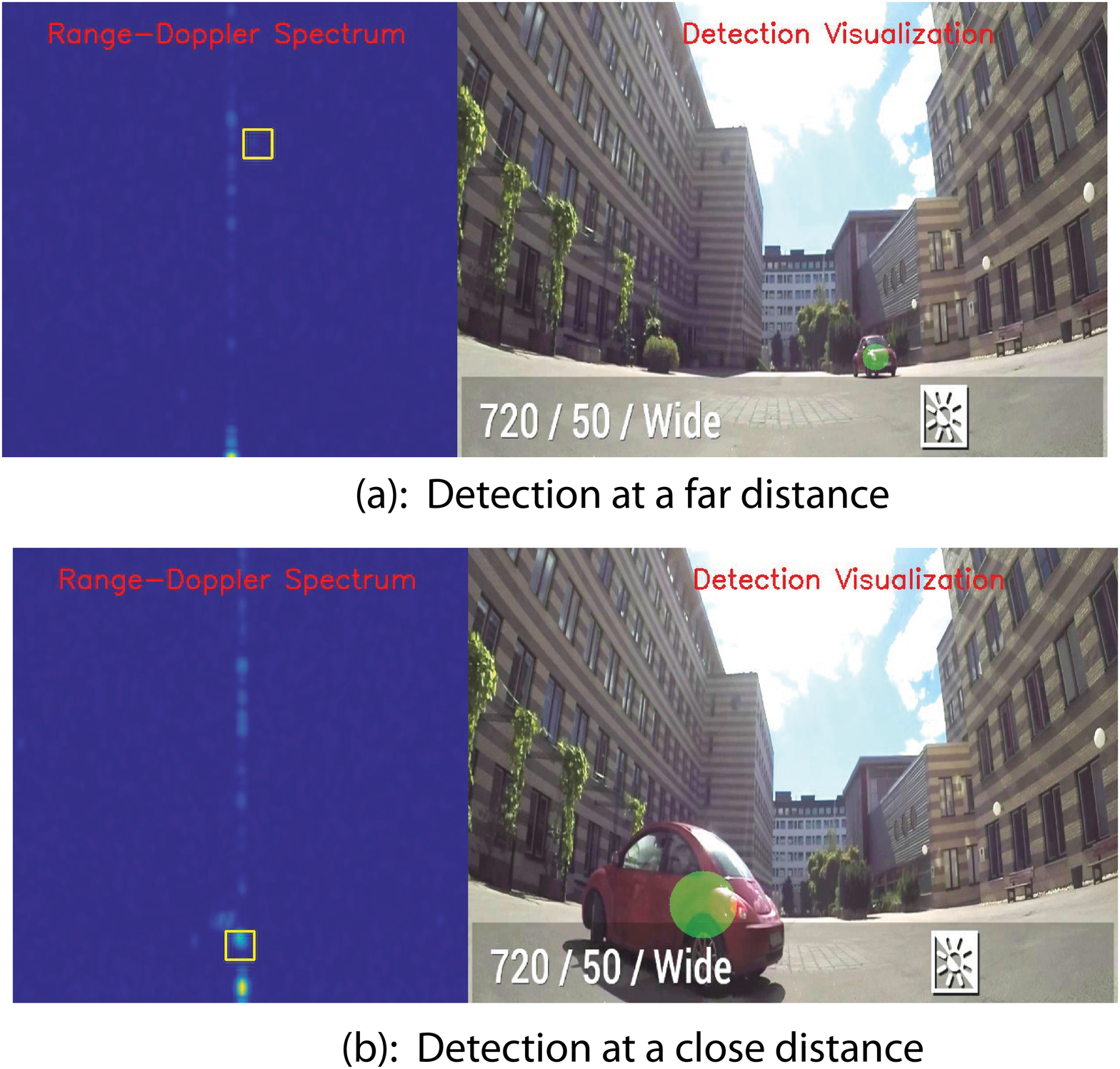}
\caption{\small Demonstration of two tested examples by applying the trained FCN on the validation dataset. The yellow box in the range-doppler spectrums indicates the cell positions for the detected car. The green circle in the images represents the estimated car orientations from the FCN. }
\label{fig:perf_demo}
\vspace{-5mm}
\end{figure}

Fig.~\ref{fig:perf_demo} displays two examples by applying the trained FCN model on the validation samples. It is clear from the figure that when the car is close to the radar, its signal on the range-doppler spectrum has a strong magnitude and occupies a reasonable number of range-doppler cells, making it easy for detection and 3D estimation. The detection becomes less easy when the car moves away from the device due to both background noise and fewer number of  range-doppler cells being occupied by the car. As shown in the figure, our proposed system is able to detect the car accurately even when the distance is large.

\vspace{-3mm}
\section{Conclusions}
\vspace{-1mm}

In this paper, we have proposed an FCN-based object detection and 3D estimation system using an FMCW radar.  A  camera has been used to assist the radar device by annotating the radar signals through image analysis. Our method requires no calibration between radar and camera coordinates. Furthermore, we have proposed a phase-normalization method to preprocess the range-doppler spectrums, which is essential to ensure successful training of the FCN. Experimental results have verified that the new system can be well trained and  applied for detecting and estimating the 3D position of a car.    

% References should be produced using the bibtex program from suitable
% BiBTeX files (here: strings, refs, manuals). The IEEEbib.bst bibliography
% style file from IEEE produces unsorted bibliography list.
% -------------------------------------------------------------------------
\bibliographystyle{IEEEbib}

%\small{
%\bibliography{sigProcessing_orig}}

\end{document}